\documentclass[conference]{IEEEtran}
\IEEEoverridecommandlockouts
\usepackage{cite}
\usepackage{amsmath,amssymb,amsfonts}
\usepackage{algorithmic}
\usepackage{graphicx}
\usepackage{textcomp}
\usepackage{xcolor}
\setkeys{Gin}{width=\linewidth, keepaspectratio}
\def\BibTeX{{\rm B\kern-.05em{\sc i\kern-.025em b}\kern-.08em
    T\kern-.1667em\lower.7ex\hbox{E}\kern-.125emX}}
\begin{document}

\title{Monodense Deep Neural Model for Determining Item Price Elasticity}

\author{
\IEEEauthorblockN{Lakshya Garg\thanks{Corresponding author: lakshya.garg@walmart.com}}
\IEEEauthorblockA{\textit{Walmart Inc.}\\
Hoboken,NJ, USA\\
lakshya.garg@walmart.com}
\and
\IEEEauthorblockN{Sai Yaswanth}
\IEEEauthorblockA{\textit{Walmart Inc.}\\
Seattle,WA, USA\\
sai.yaswanth@walmart.com}
\and
\IEEEauthorblockN{Deep Narayan Mishra}
\IEEEauthorblockA{\textit{Walmart Inc.}\\
Sunnyvale,CA, USA\\
deep.mishra@walmart.com}
\and
\IEEEauthorblockN{Karthik Kumaran}
\IEEEauthorblockA{\textit{Walmart Inc.}\\
Sunnyvale,CA, USA\\
karthik.kumaran@walmart.com}
\and
\IEEEauthorblockN{Anupriya Sharma}
\IEEEauthorblockA{\textit{Walmart Inc.}\\
Sunnyvale,CA, USA\\
anupriya.sharma0@walmart.com}
\and
\IEEEauthorblockN{Mayank Uniyal}
\IEEEauthorblockA{\textit{Walmart Inc.}\\
Bangalore, KA, IND\\
mayank.uniyal@walmart.com}
}

\maketitle

\begingroup\renewcommand\thefootnote{}\footnotetext{\scriptsize \copyright~2026 IEEE.}\endgroup

\begin{abstract}
Item Price Elasticity is used to quantify the responsiveness of consumer demand to changes in item prices, enabling businesses to create pricing strategies and optimize revenue management. Sectors such as store retail, e-commerce, and consumer goods rely on elasticity information derived from historical sales and pricing data. This elasticity provides an understanding of purchasing behavior across different items, consumer discount sensitivity, and demand elastic departments. This information is particularly valuable for competitive markets and resource-constrained businesses decision making which aims to maximize profitability and market share. Price elasticity also uncovers historical shifts in consumer responsiveness over time. In this paper, we model item-level price elasticity using large-scale transactional datasets, by proposing a novel elasticity estimation framework which has the capability to work in an absence of treatment control setting. We test this framework by using Machine learning based algorithms listed below, including our newly proposed Monodense deep neural network. 
\begin{enumerate}
    \item \textbf{Monodense-DL network} -- Hybrid neural network architecture combining embedding, dense, and Monodense layers
    \item \textbf{DML} -- Double machine learning setting using regression models 
    \item \textbf{LGBM} -- Light Gradient Boosting Model
\end{enumerate}
 We evaluate our model on multi-category retail data spanning millions of transactions using a back testing framework. Experimental results demonstrate the superiority of our proposed neural network model within the framework compared to other prevalent ML based methods listed above.
\end{abstract}


\section{Introduction}
Price elasticity of demand measures sensitivity of consumer demand to changes in item prices. To make economic and business sense this elasticity of demand measure will be always negative.
\begin{equation}
E_d = \frac{\%\ \text{change in quantity demanded}}{\%\ \text{change in price}}
\end{equation}
Accurate estimation of this plays a critical role in enabling businesses to optimize pricing strategies, forecast revenue, and design effective promotional campaigns. Industries such as retail, e-commerce, and consumer goods heavily rely on elasticity insights to understand how customers respond to price variations across different product, categories, and departments. \newline
Traditional approaches to elasticity estimation, such as econometric models, often assume linear or exponential relationships between demand and price and fail to capture complex, non-linear, and non exponential patterns. These methods also typically overlook contextual factors such as seasonality, substitute-item interactions, competitor pricing, and dynamic shifts in demand, which are increasingly important in today’s competitive and data-rich environments. Moreover these traditional econometric approaches fail to scale across millions of items due to price and business constraints of running control/treatment (C/T) experiments across the whole item universe. For retailers with millions of items sold across thousands of stores, conducting experiments for every item-store combination is financially infeasible. Additionally, the commitment to providing consistent lowest prices within each channel creates a fundamental challenge: offering two different prices for the same item in the same store simultaneously undermines the experimental design required for C/T testing. This dual challenge—catalog scalability and single-price-per-channel constraints—affects not only large-scale retailers but also small independent retailers operating even a single location. 

Recent advancements in machine learning (ML) and deep learning (DL) offer promising alternatives to conventional models by leveraging large-scale transactional data and uncovering intricate patterns in demand responsiveness. In this work, we propose the following:
\begin{enumerate}
    \item A novel framework to model item elasticity by using utilizing the rich transactional data that is universally available to any retailer.
    \item Innovative deep learning model (DLM) that can be incorporated within the framework to solve the item price elasticity problem
\end{enumerate}

\begin{figure}[htbp]
\centerline{\includegraphics{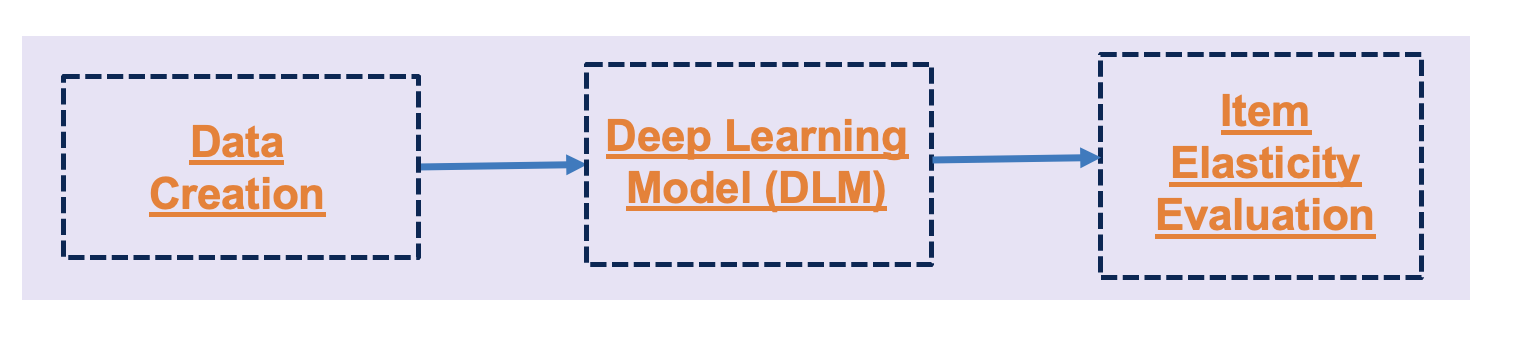}}
\caption{Proposed framework}
\label{fig}
\end{figure}

As seen in figure 1 we start by data creation, creating a cross join dataset of the aggregated monthly transaction information thereby creating a lag month and a lead month combination for each item pair. We track the prices across these lead and lag month combination along with multiple other signals across both the lead and lag months such as item inventory, substitute item availability, item out of stock (OOS) days, item rating count, days launched, competitor price, and etc. Once we create this information rich feature set we pass it to through our  proposed DLM to predict the items demand in the lead month wrt to the items price in the lead month. Finally we use the DLM to evaluate demand at different prices to get item elasticities (refer equation 3). 
Our proposed framework has the following advantages as compared to the traditional econometric methods and current ML based methods being used to solve price elasticity.
\begin{enumerate}
    \item Scalable across millions of items at minimal price to company.
    \item No requirement for a creation C/T group or the need to run recurrent expensive experiments.
    \item Our modeling approach ensures the economic relationship between price and demand is intact where as a decrease in price should result in an increase in demand and vice versa. Therefore the model imposes monotonicity between demand and price.
\end{enumerate}

Our paper consist of the following sections: In section-II we review the relevant methods present to solve the problem of item elasticity, focusing on econometric and ML methods. In section-III, we present the proposed framework, including the network architecture, training details, data creation, and elasticity evaluation. In section-IV we showcase our results and findings henceforth followed by a conclusion in section-V.

\section{RELATED WORK}
The literature on price elasticity estimation can be broadly classified into two categories: econometric-based and machine-learning-based methods. \newline
The econometric approach (e.g., traditional regression models, instrumental variable techniques) focuses on estimating elasticity using statistical models grounded in economic theory, whereas the machine-learning approach (e.g., gradient boosting, neural networks) leverages predictive algorithms to infer elasticity from observed transaction data. Econometric methods typically rely on parametric assumptions and interpretability, while ML-based methods emphasize flexibility and pattern recognition in complex datasets. Despite these differences, both approaches currently require the creation of treatment and control groups or a carefully catered dataset for training the algorithms—often through randomized experiments or by creating two datasets capturing demand of the same item at two different price points. As demonstrated by Hua et al., 2021 [1], where the same item was listed on two different marketplaces at different prices.

\subsection{Econometric based methods}
Houthakker et al., 1970 [2] pioneered the use of demand system models to estimate price elasticity of consumer goods, employing linear expenditure systems grounded in utility theory. Subsequent studies, such as Deaton et al., 1980 [3], introduced the Almost Ideal Demand System (AIDS), which allowed flexible substitution patterns and provided robust elasticity estimates across multiple commodity groups. These models typically relied on household survey data and aggregate consumption statistics, applying maximum likelihood and generalized least squares techniques for elasticity estimation. Similarly, Hausman et al., 1996 [4] utilized instrumental variable approaches to address endogeneity in price and quantity data, improving the reliability of elasticity estimates in discrete choice frameworks. Traditional econometric methods have thus formed the foundation for empirical demand analysis, emphasizing functional form assumptions and statistical consistency in elasticity measurement. \newline
It should be noted that most econometric approaches, such as the Linear Expenditure System or the Almost Ideal Demand System (AIDS), assume a specific mathematical relationship between price, quantity, and income (e.g., linear or log-linear forms). If the true demand relationship deviates from these assumptions, which is often the case in real life demand scenarios, the elasticity estimates can be biased or misleading.

\subsection{Machine-learning based methods}
Zhang et al., 2023 [5] explored gradient boosting frameworks for estimating price elasticity in retail demand forecasting, leveraging tree-based ensembles to capture nonlinear relationships between price and quantity. A feature-rich approach was adopted, incorporating historical sales, promotional indicators, and seasonality effects to improve predictive accuracy. More recently, Safonov et al., 2024 [6] utilized deep neural networks to model consumer price responses, employing feed-forward architectures trained on simulated transactional data. These models were fine-tuned using regularization techniques to mitigate overfitting and enhance generalization. Similarly, Gupta et al., 2020 [7] applied recurrent neural networks (RNNs) and attention mechanisms to airline pricing environments, enabling elasticity estimation under temporal dependencies and demand shocks. Unlike traditional econometric models, these machine learning approaches do not impose restrictive functional form assumptions and can scale effectively to high-dimensional datasets, making them suitable for personalized pricing and real-time optimization, however as often seen in real world transactional data, price and demand fail to uphold a monotonic relationship, wherein a decrease in item price does not result in increase of demand and vice versa. As the price–demand relationship is often non-monotonic across most training instances, the model frequently infers positive item elasticity values. This non monotonic relationship between price and demand usually happens due to seasonal demand variations, or consumer preferences for certain products. Our proposed Monodense DLM ensures that the final evaluated item elasticities are always negative, thereby making them economically consistent, as it bakes in the monotonicity between price and demand while modeling the demand to price relationship.

\section{FRAMEWORK}
In the following section we go over our proposed Framework.

\subsection{Dataset creation}
As discussed earlier post creating a cross join dataset of the monthly transaction information we bring in the following additional information for each lead and lag month. 
\begin{enumerate}
    \item Inventory data to account for stock-outs and availability constraints, ensuring elasticity estimates reflected true consumer behavior rather than supply limitations.
    \item Item attribute information, including brand, size, and category, subcategory to model heterogeneity across products and identify cross-category effects.
    \item Seasonal event data, such as holidays and promotional periods, to adjust for temporal demand fluctuations and isolate price-driven changes.
    \item Competitor item pricing and Item substitute to capture market dynamics and substitution effects
\end{enumerate}
The data creation is done using pyspark which leverages spark distributed computing. 2 years and 3 month of data is considered while creating the training dataset. Our model is trained on 2 years worth of data (80\% used in training and 20\% as validation set) and the 3 months of data is kept as a out of time test set which is subsequently used to test model performance metrics such as WMAPE. Figure 2 shows an example snapshot of how our final data looks like.

\begin{figure}[htbp]
\centerline{\includegraphics{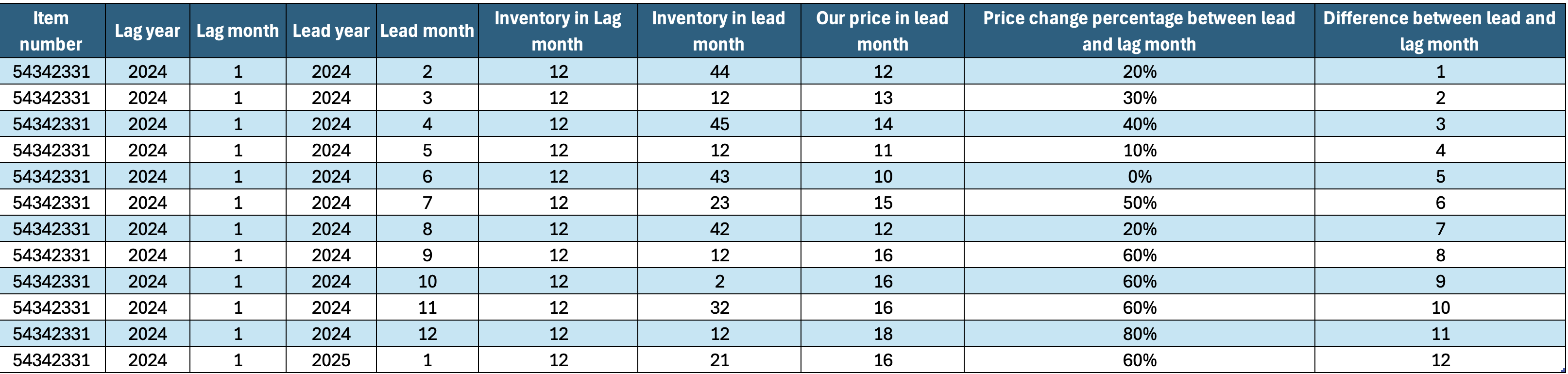}}
\caption{For simplicity some features are not shown}
\label{fig}
\end{figure}

Figure 2 shows only one cross join combination where lag month is one. In the final training set we have multiple combinations where lag month can be 1,2,3,4,...12 and lead months starts from 1,2,3,4,5...12 and so on. We strictly ensure the below constraints while creating the cross join data.
\begin{enumerate}
    \item The year month difference between lead and lag year months cannot exceed more than 12 months and cannot be negative. Therefore the acceptable range of month difference between lead and lag month is between 1 to 12.
    \item We only consider those combinations as valid where item inventory is greater than zero in both lead and lag months.
\end{enumerate}

\subsection{Proposed Monodense-DLM network}
Creation of data is the first step in our framework. Post creating the data we pass the lead lag month combination pair data to our proposed DLM.
Our neural network architecture is designed for demand prediction at different prices and incorporates both product and item embedding layers alongside dense layers to capture complex feature interactions. The main purpose is to enable the model to learn the following mathematical relationship $D_{i} = f(P_{i}, X_{i}, S_{i})$ where $D_{i}$ represents the units sold for the item in the lead month, $P_{i}$ denotes the price of the item in the lead month, $ X_{i}$ represents the item, seasonal, and environment variables, and $S_{i}$ represents price change percentage between lead and lag months.  \newline
The model (figure 3) starts with an encoder like structure where each input depending on whether its a categorical or a continuous variable is passed through an embedding or a feed forward dense layer. Each embedding layer is followed by a flatten layer, and each dense layer is followed by a Relu activation. Post learning the higher dimensions of the input features the embedding and continuous feature outputs are further concatenated together and passed through multiple dense layers to further learn higher dimension feature interactions. We saw in our experiments when price was passed in the top of the proposed model, the models sensitivity towards the price was being lost, thereby to make the model more price sensitive we introduced the item price at the lower layers of the model through monodense layer. This monodense layer ensures that in the learned price to demand relationship a decrease in price results in an increase in demand and vice versa. \newline

For our implementation of the monodense layer we follow the work proposed by Davor et al., 2023 [8]
\begin{itemize}
    \item We apply weight constraints based on a monotonicity indicator vector \textbf{t}, where each element $t_i$ corresponds to a feature $x_i$.
    \begin{itemize}
        \item If $t_i = 1$ The feature is considered to be \textbf{monotonically increasing} so an increase in input X would result an increase in output Y . The corresponding weights ($w_i$) are constrained to be non-negative ($w_i \ge 0$).
        \item If $t_i = -1$: The feature is considered to be \textbf{monotonically decreasing} so an increase in input X would result a decrease in output Y. The corresponding weights ($w_i$) are constrained to be non-positive ($w_i \le 0$). This is implemented by taking the negative of a non-negative weight.        
        \item If $t_i = 0$: The feature has \textbf{no monotonic constraint}. The corresponding weights ($w_i$) are not constrained and can be positive, negative, or zero.
    \end{itemize}
    \item To correctly capture the both concave and convex nature of demand to price relationship as proposed by Davor et al., 2023 [8] we start with a zero-centered, monotonically increasing, convex activation function ($\rho$) (e.g., ReLU, ELU, SELU). Then Construct two additional monotonic variants:
    \begin{itemize}
    	\item \textbf{Concave upper-bounded function:} \[ \hat{\rho}(x) = -\rho(-x) \]
	\item \textbf{Bounded function:} 
	\begin{equation}
		\tilde{\rho}(x) =
		\begin{cases}
		\check{\rho}(x + 1) - \check{\rho}(1), & \text{if } x < 0 \\
		\hat{\rho}(x - 1) + \check{\rho}(1), & \text{{otherwise}}
	\end{cases}
	\end{equation}
	Defined piecewise to ensure saturation at its bounds.
    \end{itemize}
    \item Finally we apply these three activations across different neuron subsets in the monodense layer:
        \begin{itemize}
    		\item Original convex activation ($\rho$)
		\item Concave activation ($\hat{\rho}$)
		\item Bounded activation ($\tilde{\rho}$)
 	\end{itemize}      
\end{itemize}

\begin{figure}[htbp]
\centerline{\includegraphics{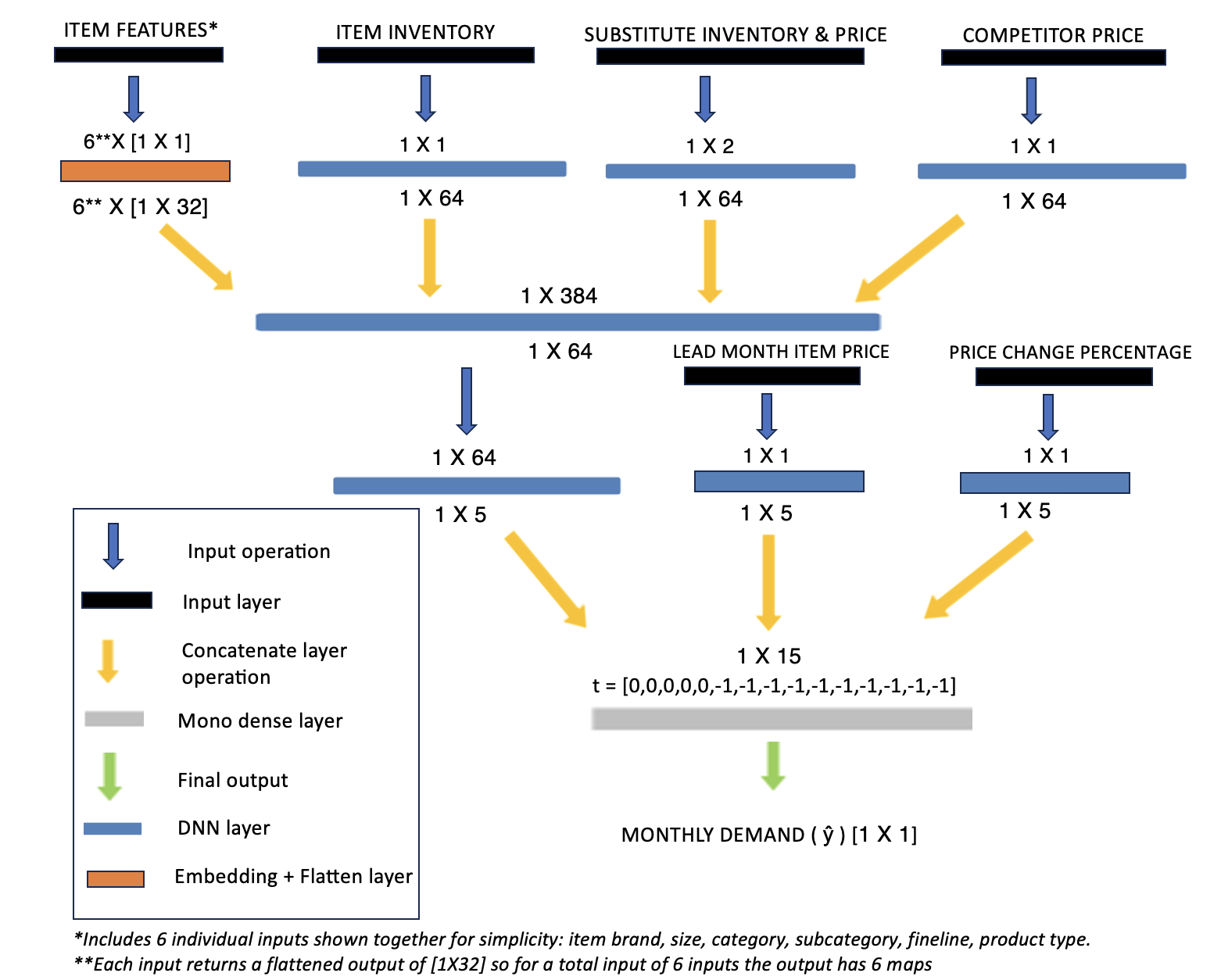}}
\caption{Architecture of our proposed DLM}
\label{fig}
\end{figure}

Our DLM network is inherently different from the traditional machine learning/deep learning based methods for solving price elasticity in the following ways.
\begin{enumerate}
    \item A key innovation in our architecture is the inclusion of a monodense layer, which enforces monotonicity of price with respect to demand, ensuring economically consistent predictions.
    \item Unlike traditional feed-forward models mainly consist of only dense layers. Our network learns jointly from both embeddings and dense representations, enabling it to model for both categorical and continuous attributes effectively while solve the price elasticity problem.
    \item To enhance price sensitivity, the price input is introduced at the lower layers of the network, allowing higher price sensitivity of the model and improving responsiveness to price changes
\end{enumerate}

\subsection{Training Details}
For training the model, the mean squared error (MSE) loss function is used. This loss minimizes the error between the model-predicted demand for the lead month and the actual observed demand in the lead month. Given a continuous demand target \( y_{i} \) and model prediction \( \hat{y}_{i} \), the loss is computed as:

\[
\mathcal{L}_{\text{MSE}} = \frac{1}{N} \sum_{i=1}^{N} \big( y_{i} - \hat{y}_{i} \big)^{2}
\]

where \( N \) denotes the number of samples in the batch. The network outputs a single continuous value per item-instance \( \hat{y}_{i} \), representing predicted demand. The input consists of a feature vector \( P_{i}, X_{i}, S_{i} \in \mathbb{R}^{F} \) for each observation \( i \), which includes price, promotion indicators, inventory signals, competitor prices, and seasonal features.
The network is trained using the Adam optimizer for 25 epochs. A batch size of 128 is used along with a learning rate of 0.01. To stabilize training and improve generalization, inputs are standardized to zero mean and unit variance, and L2 weight decay is included in dense layers. Model performance is monitored through tracking training/validation loss and MSE. For training the model a GPU with CUDA Core 3840 was used. We tested our model on 32 Core, 2.10 GHz CPU.

\begin{figure}[htbp]
\centerline{\includegraphics{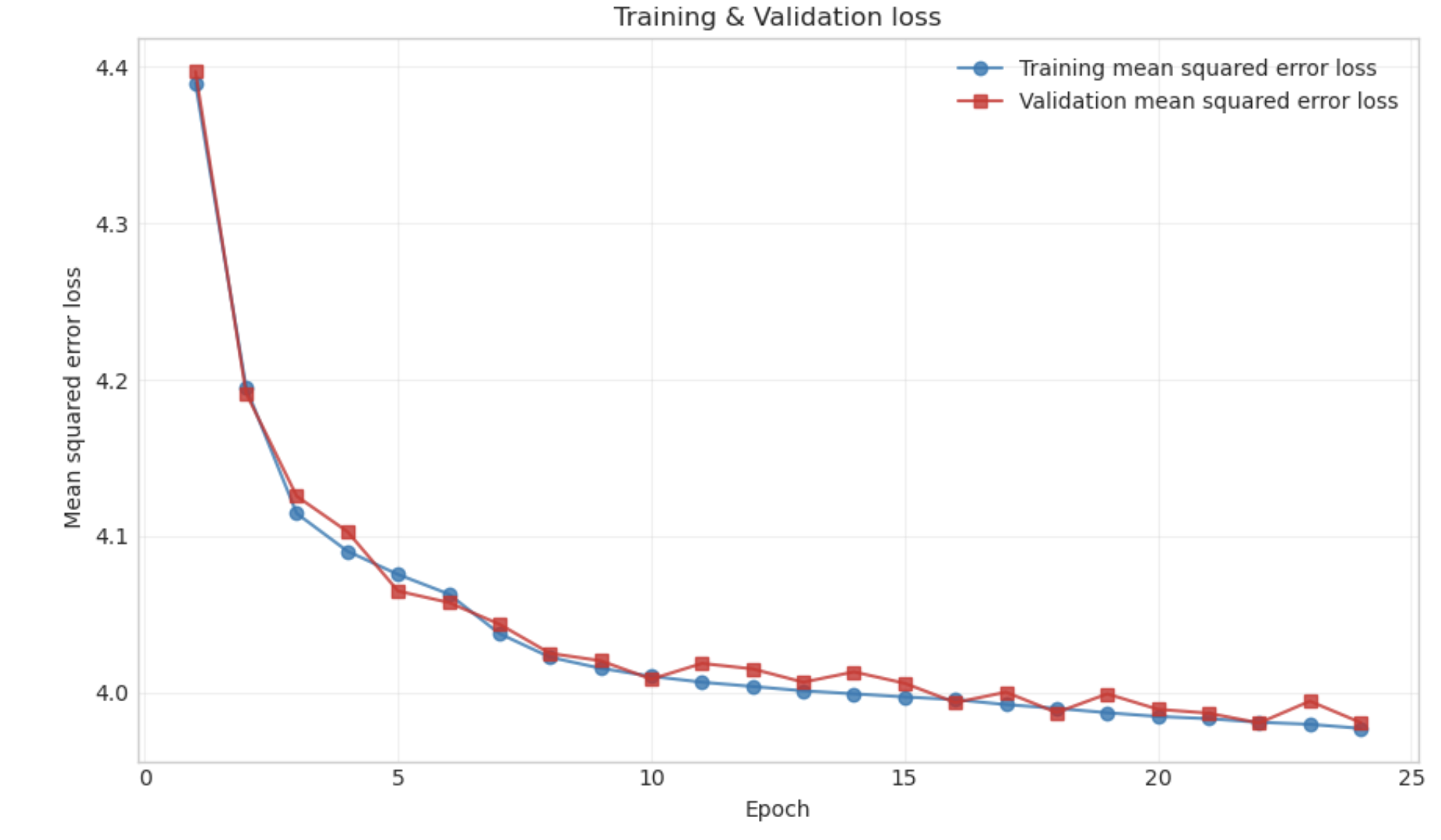}}
\caption{Model training and validation loss}
\label{fig}
\end{figure}

\subsection{Item elasticity evaluation}
The model returns item demand in the projected month with respect to an input price. Therefore, the price elasticity for any price change $\Delta p$ can be evaluated as:
\begin{equation}
\mathcal{E}_{\Delta p} = \frac{\hat{y}(p + \Delta p) - \hat{y}(p)}{\hat{y}(p)} \times \frac{p}{\Delta p}
\end{equation}

Here, \(\Delta p\) represents the price change that can be submitted by the user or predefined as per business requirements. In practice, to evaluate elasticity for a future month, we create our inference set consisting of the most recent lag month data for the item and consider the lead month as \(\text{lag month} + 1\). Then, we predict counterfactual demands in the lead month at two prices \(p + \Delta p\) and \(p\) to evaluate elasticity in the lead month for a price change of \(\Delta p\) leveraging equation number 3.

\section{RESULTS AND FINDINGS}
The result evaluation is done through two metrics
\begin{enumerate}
    \item Calculating the actual demand weighted mean absolute percentage error (WMAPE) between predicted and actual demand on the test set.
    \item By calculating the mean absolute error (MAE) between model evaluated elasticities and actual observed elasticities across a sample set of items.
\end{enumerate}

The WMAPE is assessed by comparing the ground truth demand and the model predicted demand across an out of time test set (OTS).  To create this OTS we use last 3 months of the overall data. The model has never been exposed to this set while training.

\[
\begin{aligned}
\text{WMAPE} &= \frac{\sum_{i=1}^{n} y_i \cdot |y_i - \hat{y}_i|}{\sum_{i=1}^{n} y_i} \times 100\%, \\
\text{where } &\hat{y}_i = \text{model predicted demand}, \\
& y_i = \text{actual demand}.
\end{aligned}
\]

 A systematic WMAPE comparison of our proposed model is done with LGBM and a double machine learning regression based model. LGBM model consists of multiple gradient boosting trees with the objective to minimize the gradient of demand in the lead month. We have leveraged LGBMs (Guolin et al., 2017 [9]) inbuilt monotonicity indicator to impose the monotonicity between demand and price. For double machine learning (DML) (Victor et al., 2018 [10]) we have trained a regression based outcome, treatment, and residual model. As regression models have no inherent way of imposing a price to demand monotonic relationship, we do observe some positive item elasticities. Both the models were implemented within the same proposed framework and therefore use the same underlying data distributions. \newline

Table 1 shows the comparison of all three models using the WMAPE evaluation criterion. The model having the least demand WMAPE is considered the best. As better demand prediction will in turn lead to a more reliable elasticity metric because elasticity is evaluated as difference of demands vis a vis equation 3. 

\begin{table}[h!]
\centering
\begin{tabular}{|c|c|c|}
\hline
\textbf{Deep learning model} & \textbf{LGBM model} & \textbf{Double Machine Learning*} \\
\hline
30.90\% & 35.9\% & 36.1\% \\
\hline
\end{tabular}
\caption{*For double machine learning WMAPE of outcome model is recorded}
\label{tab:WMAPE Comparison}
\end{table}

A zero WMAPE would signify that the model predicted demand values and actual observed demand values at different price points are in perfect sync while higher values would represent lower agreement.  It is observed that our proposed model outperforms the other two models in terms of overall WMAPE.

We are aware that WMAPE is not the only criterion to judge our model as our final goal is to evaluate elasticities so for this we calculate the MAE between a set of items for which we have a known elasticity referred as ground truth elasticities going forward. MAE measures the deviation from these ground truth elasticities and the model evaluated elasticities below is the formula for the same.

\[
\begin{aligned}
\text{MAE} &= \frac{1}{n} \sum_{i=1}^{n} \left| e_i - \hat{e}_i \right|, \\
\text{where } &\hat{e}_i = \text{predicted elasticity (model evaluated output)}, \\
& e_i = \text{actual ground truth elasticity}.
\end{aligned}
\]

\begin{table}[h!]
\centering
\begin{tabular}{|c|c|c|}
\hline
\textbf{Deep learning model} & \textbf{LGBM } & \textbf{Double machine learning} \\
\hline
0.36 & 0.42 & 0.43 \\
\hline
\end{tabular}
\caption{MAE comparison across models}
\label{tab:MAE Comparison}
\end{table}

The model having the least MAE is considered to have the best elasticities. Ideally a zero MAE would signify that the model predicted elasticities perfectly match the ground truth elasticities whereas a higher number would represent lower agreement between elasticities. When comparing our proposed Monodense DL model with other two models the MAE is lower for our model.

\section{CONCLUSION}
This paper has showcased that it is possible to use deep neural networks for evaluating item elasticity while making economically consistent predictions. We successfully evaluate item elasticity using a newly proposed Monodense deep neural network (DNN) architecture. We benchmark Monodense DNN model with other prevalent elasticity modeling techniques which are possible to run within our proposed framework such as LGBM and double machine learning methods. The experimental results demonstrate that our novel model outperforms double machine learning and LGBM model. This can be said by observing that Monodense DLM achieves a lower overall WMAPE and MAE value than the other two models. \newline
Paper also demonstrates that with our proposed data creation method it is possible to model the price elasticity using a huge real world transaction dataset (1 billion+ rows) counterpart to the other studies conducted in the area which have only utilized comparatively small or synthetically created datasets. We also display that there is no need to artificially impose demand to price monotonic relation in the underlying training data through data filtering or to subset specific items to run C/T experiment, which might have been following the correct demand to price relationship. This monotonic relationship can be baked in the DNN model itself. \newline
As future directions of our work we plan to experiment with an even deeper model architecture, and experiment with different activation functions to further better model the demand to price relationship thereby decreasing our WMAPE and MAE even further.

\end{document}